\documentclass{article}
\usepackage{iclr2027_conference,times}
\iclrfinalcopy
\usepackage{amsmath,amssymb,graphicx,booktabs,needspace,float}
\usepackage[caption=false,font=small]{subfig}
\usepackage{microtype,xcolor}
\usepackage{hyperref}
\usepackage{url}
\hypersetup{hidelinks,pdftitle={FINGR: Learning Dexterous Hand Control for Real-World Rubik’s Cube Solving},pdfauthor={Yutong Liang, Quanquan Peng, Matthew Kim, Xiaolong Wang}}
\newcommand{\ours}{\textsc{Fingr}}
\title{\ours: Learning Dexterous Hand Control for Real-World Rubik’s Cube Solving}
\author{Yutong Liang$^{1,*}$ \quad Quanquan Peng$^{1,*}$ \quad Matthew Kim$^{1,*}$ \quad Xiaolong Wang$^1$\\
\normalfont $^1$University of California San Diego\\
\normalfont $^*$Equal contribution.}
\newcommand{\thcell}[1]{\multicolumn{1}{c}{#1}}

\newcommand{\R}{\mathbb{R}}
\newcommand{\meanstd}[2]{#1\,\mathord{\pm}\,#2}
\makeatletter
\let\tableinput\@@input
\makeatother
\begin{document}
\maketitle
\fancyhead{}
\begin{abstract}
  Manipulating a Rubik's Cube with one single dexterous hand is a challenging test of sustained, contact-rich control: the hand must execute successive layer turns while keeping the cube secure. Each turn requires some fingers to support the cube while others push a moving layer, release contact, and reset for the next move.
  To learn this coordination, we introduce \textbf{\ours{}} (\textbf{F}uture-supervised \textbf{I}nteraction \textbf{N}etwork with \textbf{G}eometric \textbf{R}epresentations), a policy that combines finger-relative geometry with future interaction prediction. A shared point encoder expresses the cube relative to each fingertip and aggregates its points without depending on cubie indexing. Learned future tokens share the observation encoder and receive supervision for contact-force changes, layer-turn progress, and finger joint displacement at multiple time scales. The resulting representation conditions a flow policy that directly generates finger actions. On a real dexterous hand, our policy achieves around $99.0\%$ success over 300 turn attempts, compared with $79.7\%$ for the base flow policy. Integrated with grasping and table-assisted regrasping, the policy solves all ten scrambled $2\times2\times2$ cubes in a mean complete-system time of approximately $137$ seconds.
  Please visit our \href{https://www.lyt0112.com/projects/FINGR}{\textcolor{blue}{project website}}.
\end{abstract}
\begin{figure}[!h]
  \centering
  \includegraphics[width=\linewidth]{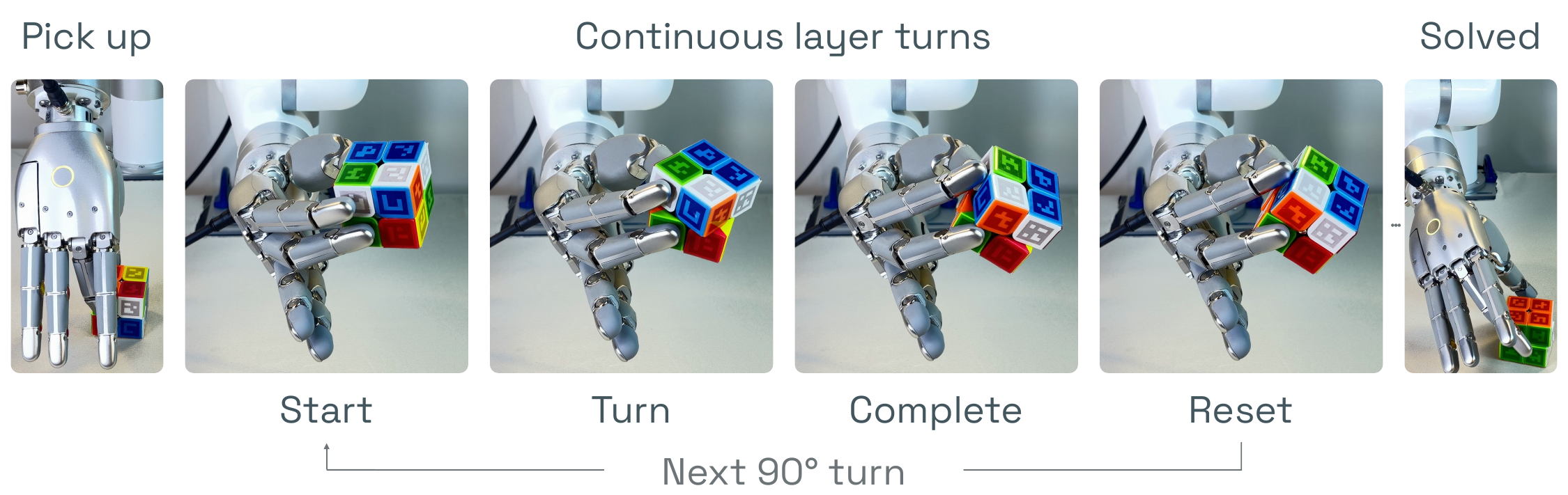}
  \caption{\textbf{Real cube solving through continuous layer turns.} The four central frames show one $90^\circ$ layer turn, from the starting grasp through finger reset. Repeating these learned motions, together with grasping and table-assisted regrasping, completes cube solving.}
  \label{fig:teaser}
\end{figure}
\section{Introduction}
\label{sec:introduction}
Human cube solvers use finger tricks to turn a layer with a short finger motion, recover the finger, and immediately continue with the next move. Reproducing this behavior with a single hand requires coordination across fingers with different roles. Some fingers must support the cube, while others establish contact with the moving layer, push it through a quarter turn, and recover a configuration suitable for the next action. A successful isolated push is therefore only one part of the skill: the resulting hand--cube configuration must also support continued manipulation.

This coordination has both spatial and temporal structure. Spatially, a finger's available action depends on where the moving layer and supporting surfaces lie relative to it. Encoding the cube globally and each finger separately leaves the policy to infer these relationships. Temporally, contact changes during a turn: a finger loads a surface, advances the layer, releases contact, and moves back. Action imitation alone does not explicitly train the observation encoder to capture how contact and motion will evolve.

We address these two aspects with \textbf{\ours}. First, we translate the cube's geometric points to origins at the five fingertips. A finger-relative cube position encoder and symmetric aggregation produce a geometric feature for each finger, which is added directly to its state token. The representation preserves information about relative directions and distances while remaining invariant to the order of cube points. Second, we introduce learned future tokens at three horizons. These tokens interact with the observation tokens in a transformer encoder and are trained to predict contact-force changes, target-layer progress, and finger joint displacement. The predictive targets connect the representation to the physical events needed for turning and recovery.

The two components are trained with a flow policy that outputs joint-action chunks. Future-token prediction provides auxiliary supervision during learning, while the encoded tokens directly condition action generation at deployment. We implement the policy on a seven-joint arm with a dexterous hand. The thumb and middle finger retain the cube, and the index, ring, and little fingers execute the learned motion. A discrete solver composes upper- and left-layer turns with a planned table-assisted regrasp when a new grasp is required to complete the solution.

Our contributions are a real robotic cube-solving system based on continuous layer turns; a permutation-invariant geometric encoding aligned with individual fingertips; and multi-horizon future-token prediction that jointly supervises contact, layer progress, and finger motion. Across 300 physical layer-turn attempts per method, local geometry increases success from $79.7\%$ to $92.7\%$, and future-token prediction further increases it to $99.0\%$. The complete system solves ten randomly scrambled cubes in an average of approximately 137 seconds, including pickup, regrasping, placement, and return to the starting robot posture.

\section{Related Work}
\label{sec:related}
\paragraph{Dexterous manipulation of articulated objects.}
Prior work on in-hand manipulation has learned object rotation and reorientation using proprioceptive, tactile, and visual feedback~\citep{rapidrotation,touchrotation,rotateit}. Subsequent work extends rotation to slender objects and composes rotation skills for goal-directed reorientation~\citep{penspin,dexhier,yin2025dexteritygen,kedia2026simtoolreal}. Articulated objects additionally require control of internal configurations. Rubik's cube systems combine symbolic planning with learned manipulation to execute multi-step solutions in simulation and on physical hands~\citep{li2019cube,akkaya2019cube}. These capabilities benefit from scalable simulation and robot-learning platforms~\citep{isaacgym,roboverse,gsworld,maniskill3}, alongside datasets and capture systems that describe hand pose, object motion, and contact~\citep{grab,dextercap,arctic,contrack,contactpose}. In particular, articulated interaction data expose the coupled evolution of hand motion and object configuration. We learn continuous layer turns from real demonstrations, organizing the policy around each finger's access to the moving layer and its recovery for the next turn. Integrating this skill with pickup and table-assisted regrasping enables complete physical cube solving.

\paragraph{Hand--object representations.}
Hand--object interaction representations connect the kinematics of a hand to the geometry it manipulates. MANO provides a compact model of articulated hand shape and pose~\citep{mano}, while point-set encoders aggregate object geometry without depending on point ordering~\citep{pointnet}. Representing their relationship explicitly can further expose feasible contacts and motions. ManipNet combines global object occupancy with local distances in a hand-centric coordinate system to synthesize finger motion~\citep{manipnet}. Contact-based grasp refinement and hand--surface encodings similarly use geometric relationships to guide grasping and dexterous control~\citep{contactopt,dexrep,dexrepplus}. Representations based on contact, shared action spaces, and robot--object distances also support transfer across different hand embodiments~\citep{gendexgrasp,xlvla,dro}. These approaches motivate expressing geometry in coordinates relevant to the acting hand. For continuous cube turns, the relevant relationships change separately for each finger as the layer advances. We therefore encode directional offsets from every fingertip to the reconstructed cube points, pool over the point set, and attach each feature to its corresponding finger token. This preserves cubie-index invariance while supplying the policy with finger-specific spatial information throughout the turn.

\paragraph{Prediction for action learning.}
Prediction links an agent's current observations to the outcomes of its actions. Latent world models use learned dynamics for online planning or behavior learning through imagined trajectories~\citep{planet,dreamer,tdmpc2}. In imitation learning, generative policies produce coordinated action sequences using latent-variable models, diffusion, and flow matching~\citep{act,diffusion,flowmatching}; reconstruction and future-observation objectives can enrich the representations used for action prediction~\citep{crossway,gr1,seer,peng2026embodiment}. Recent world-action models bring these directions together by coupling action generation with predictions of the evolving visual world~\citep{uwm,worldvla,fact,dreamzero}. Across these approaches, future supervision encourages features that capture how interactions unfold. For repeated layer turns, useful future information includes whether contact is changing, whether the target layer is advancing, and how the fingers are moving toward their next configuration. Our future tokens jointly predict these three quantities at multiple horizons, linking immediate contact changes to longer-term turn progress and finger motion. The prediction objective trains the shared observation representation, which directly conditions the action head during deployment.

\section{Method}
\label{sec:method}
\subsection{Task and Policy Interface}
\label{sec:interface}
We consider a $2\times2\times2$ cube held by a dexterous robot hand. The learned skill performs $90^\circ$ turns of two layers, denoted $U$ and $L$ in the current grasp. A discrete solver selects these moves and decides when to place and regrasp the cube. During a learned turn, the arm maintains its posture and the thumb and middle finger maintain their holding configuration; the policy commands the 13 joints of the index, ring, and little fingers.

At control time $t$, the observation contains numeric features $s_t^i$ for each of the five fingers, tactile deformation images for the three controlled fingers, cube points $P_t=\{p_{t,j}\}_{j=1}^{32}$, a move label $m_t\in\{U,L\}$, and a normalized remaining angle $r_t$. Each $s_t^i$ contains joint positions, velocities and torques, six tactile force/torque channels, and the fingertip position $f_t^i$. Cube points and fingertip positions share the palm coordinate frame. The points describe the current articulated geometry through 24 face-center samples and eight outer-corner samples reconstructed from the cubie poses. The policy receives their coordinates without cubie identity or color labels.

The policy predicts an action chunk of $H=20$ joint-position offsets relative to the current controlled joint configuration $q_t\in\R^{13}$. Observations and commands run at 10 Hz. We execute at least five steps before updating the chunk and align commands to the observation timestamp to compensate for inference latency. All evaluated action-learning methods use this interface.

\begin{figure}[!h]
  \centering
  \includegraphics[width=\linewidth]{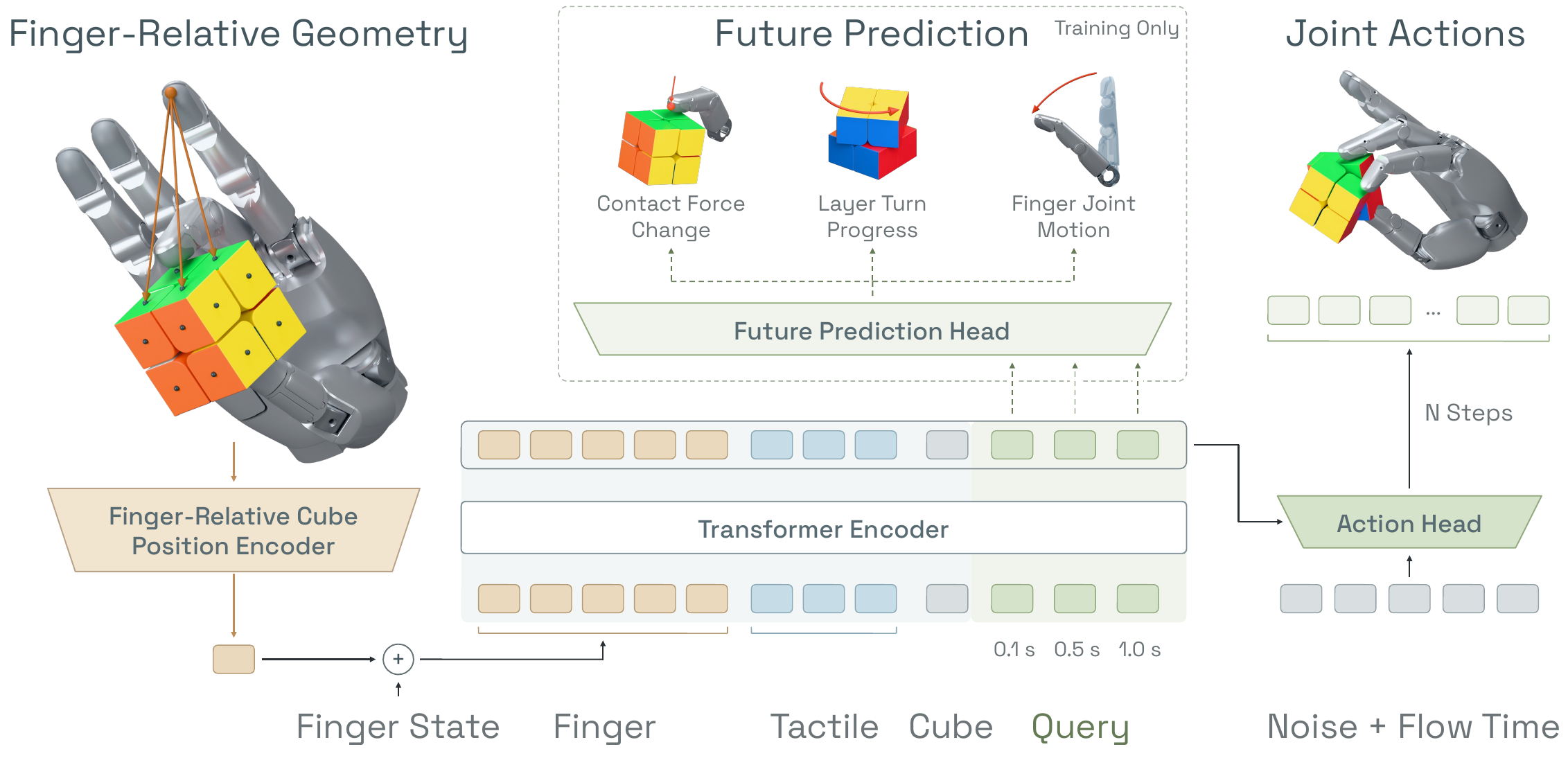}
  \caption{\textbf{Policy architecture.} \ours{} uses a finger-relative cube position encoder to augment the five finger-state tokens. A transformer encoder processes these tokens together with tactile inputs, global cube geometry, and three future queries. At each future horizon, the future prediction head jointly predicts contact-force change, turn progress, and joint displacement during training. The full encoded memory, including the three future tokens, conditions the action head at deployment.}
  \label{fig:pipeline}
\end{figure}

\Needspace{10\baselineskip}
\subsection{Finger-Relative Geometry}
\label{sec:geometry}
A global point-set feature summarizes the cube but does not explicitly associate its surfaces with a particular finger. We construct that association before the observation transformer. For finger $i$ and cube point $j$, define
\begin{equation}
  d_{t,ij}=\frac{p_{t,j}-f_t^i}{\ell},\qquad
  g_t^i=W_g\left(\frac{1}{32}\sum_{j=1}^{32}\phi(d_{t,ij})\right)+b_g,
  \label{eq:geometry}
\end{equation}
where $\ell=0.05$ m is a fixed length scale, $\phi:\R^3\rightarrow\R^{64}$ is a shared two-layer point network, and $W_g,b_g$ project its pooled output to the transformer width. The origin moves with the fingertip; the axes remain aligned with the palm. Thus, $d_{t,ij}$ retains the direction and distance from a specific finger to a cube surface point.

We add $g_t^i$ to the projected numeric feature of finger $i$. Finger identity embeddings distinguish the five fingers, and modality embeddings distinguish numeric, tactile, and global cube information. A shared convolutional network encodes the three tactile images, with finger-specific feature-wise scale and shift parameters. A separate point network with max pooling provides one global cube token. The task embedding and a projection of $r_t$ are added to all nine observation tokens.

The mean in Equation~\ref{eq:geometry} is unchanged by any permutation of the cube points, so the finger feature does not depend on cubie indexing. At the same time, translating the points to each fingertip makes the relative geometry explicit within that finger's token. The transformer can use these tokens to coordinate the fingers, while the point network already exposes the local spatial relation needed to select a push, maintain support, or recover from contact.

\subsection{Future Interaction Prediction}
\label{sec:future}
Finger-relative geometry describes where each finger can interact with the cube. To connect this geometry to the unfolding turn, we train future tokens to predict how contact, layer progress, and finger motion change together. Each learned query $z_h$ corresponds to an offset $h\in\mathcal H=\{1,5,10\}$ control steps. The transformer jointly encodes the current observation and these three queries:
\begin{equation}
  M_t=E_\theta\big([O_t,z_1,z_5,z_{10}]\big),\qquad
  \widehat y_{t,h}=D_\psi(M_t^{(h)}),
  \label{eq:future}
\end{equation}
where $O_t$ denotes the nine observation tokens, $E_\theta$ is the transformer encoder, $M_t^{(h)}$ is the encoded future token for horizon $h$, and $D_\psi$ is the future prediction head shared across horizons.

Let $c_t\in\R^9$ concatenate the three-axis contact-force measurements of the three controlled fingers. We supervise each query with
\begin{equation}
  y_{t,h}=
  \left[\,c_{t+h}-c_t,\quad r_t-r_{t+h},\quad q_{t+h}-q_t\,\right]\in\R^{23}.
  \label{eq:targets}
\end{equation}
The targets describe three coupled aspects of a turn: contact loading and unloading, progress toward the target layer angle, and measured joint displacement during pushing and recovery. Supervising them together requires each future token to relate object progress to the contact and finger motion that accompany it. Short horizons emphasize immediate changes; longer horizons cover more of the turn and reset sequence.

Targets are scaled by their empirical standard deviations separately for each horizon and coordinate. A target is valid only when its current and future frames lie in the same continuously tracked turn segment. We average squared errors within each target group and then average the three groups:
\begin{equation}
  \mathcal L_{\mathrm{pred}}=
  \frac{1}{3|\mathcal V|}\sum_{(t,h)\in\mathcal V}
  \left(\frac{\|e^c_{t,h}\|_2^2}{9}
  +|e^r_{t,h}|^2
  +\frac{\|e^q_{t,h}\|_2^2}{13}\right),
  \label{eq:predloss}
\end{equation}
where $\mathcal V$ is the set of valid current-frame/horizon pairs and $e^c,e^r,e^q$ are errors in the standardized targets. This averaging gives contact, progress, and motion equal weight despite their different dimensions.

Action learning imitates successful, well-aligned turns. Future prediction also uses continuous interaction segments excluded from action imitation, extending the supervision of contact and motion dynamics. Both objectives update the shared observation encoder. At deployment, the encoder constructs $M_t$ from the current observation and learned queries, and the action head attends to all encoded tokens. The auxiliary prediction head is omitted; the future tokens carry the learned predictive features directly into action generation.

\subsection{Action Learning and Execution}
\label{sec:action}
We use a flow policy with an action denoising parameterization. Let $A_t\in\R^{H\times13}$ be the normalized demonstrated action chunk and $\epsilon\sim\mathcal N(0,I)$ a noise chunk. For noise level $\tau\in(0,1]$, the training input is
\begin{equation}
  X_\tau=(1-\tau)A_t+\tau\epsilon,\qquad
  v_\theta(X_\tau,\tau,M_t)=\frac{X_\tau-\widehat A_\theta(X_\tau,\tau,M_t)}{\tau}.
  \label{eq:flow}
\end{equation}
The action head alternates action self-attention, cross-attention to $M_t$, and feed-forward layers. Noise-level conditioning modulates each block. We train the clean-action estimate $\widehat A_\theta$ with masked joint-space reconstruction error; Appendix~\ref{app:objective} gives the exact normalization. At inference, four Euler steps integrate Equation~\ref{eq:flow} from noise at $\tau=1$ to the action chunk at $\tau=0$.

The combined training objective is
\begin{equation}
  \mathcal L(k)=\mathcal L_{\mathrm{act}}+\lambda(k)\mathcal L_{\mathrm{pred}},\qquad
  \lambda(k)=0.003\max\left(0,1-\frac{k}{0.5K}\right),
  \label{eq:total}
\end{equation}
where $k$ is the optimizer step and $K$ is the total number of training steps. Predictive supervision shapes the representation during the first half of training, after which the action objective continues to optimize the same transformer encoder and action head. Deployment uses exponential-moving-average model parameters.

\subsection{Complete Cube-Solving System}
\label{sec:system}
The system first reconstructs the cube pose, plans a pregrasp, closes the supporting fingers, and lifts the cube into the manipulation pose. Multi-view tracking of coded surface markers estimates cubie poses. After pickup, the system observes the discrete cube state and computes a sequence of $U/L$ moves. The solver minimizes the number of positive quarter turns and then the cost of repeated turns of the same layer.

Some scrambles require changing the supporting grasp. The planner then first creates a compatible solved two-cubie block, places the cube on the table in a selected orientation, and grasps that block before completing a second $U/L$ sequence. Placement and regrasping use kinematic planning and collision checks. The system verifies the cube after a sequence, performs an additional alignment motion when needed, and replans from the observed state. Once all cubies have a common solved orientation, the robot places the cube on the table and returns to its starting posture. This connects the learned continuous manipulation skill to complete solutions of scrambled cubes.

\section{Experiments}
\label{sec:experiments}
We evaluate whether the proposed representation improves physical layer turning and whether this skill supports complete cube solving. The first experiment compares action-learning methods under a common observation and deployment interface. The second isolates local geometry and future-token prediction. The final experiment evaluates the complete system on ten random scrambles.

\subsection{Experimental Setup}
\label{sec:setup}
\begin{figure}[t]
  \centering
  \subfloat[$U$ turn\label{fig:training_U}]{\includegraphics[width=\linewidth]{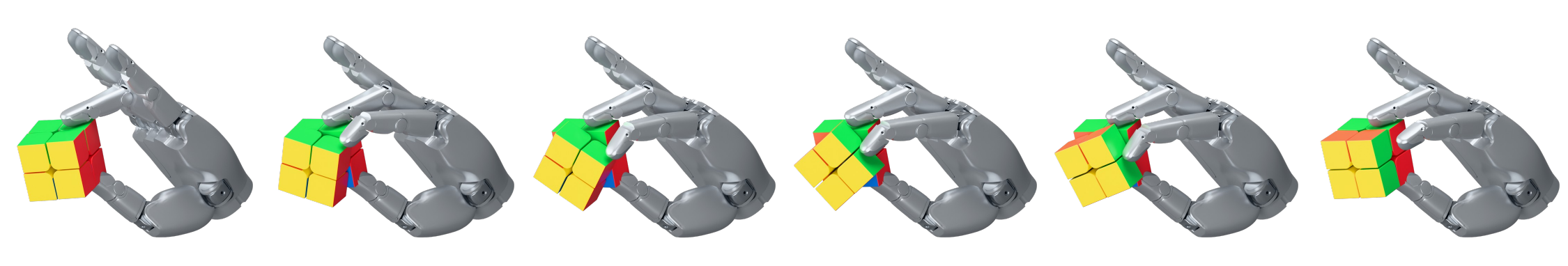}}
  \par\vspace{0.5em}
  \subfloat[$L$ turn\label{fig:training_L}]{\includegraphics[width=\linewidth]{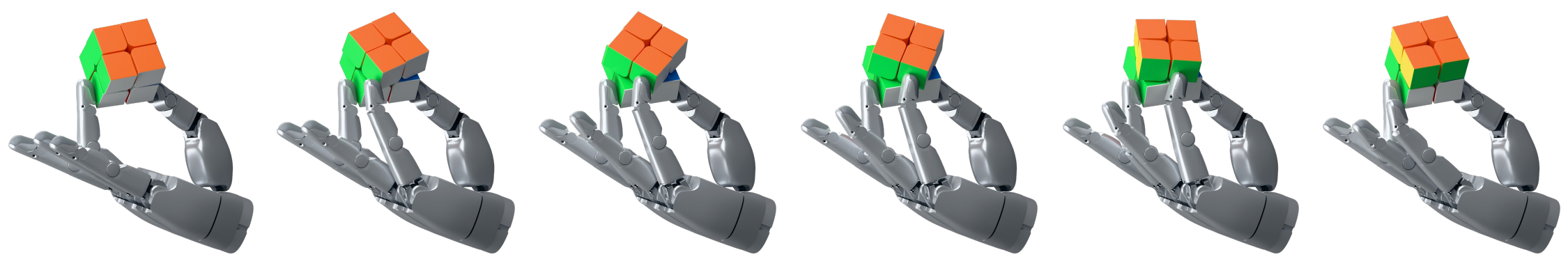}}
  \caption{\textbf{Layer-turn demonstrations.} Each row shows six hand--cube configurations at approximately $18^\circ$ increments across one $90^\circ$ turn.}
  \label{fig:training}
\end{figure}

\paragraph{Hardware and demonstrations.}
The platform uses an xArm7 and a 22-joint Sharpa hand. The cube has a nominal edge length of 52 mm and coded markers for reconstruction. Our demonstration set comprises 36 teleoperated sequences with 576 turns. Action training uses 435 successful turns of at most eight seconds: 226 $U$ turns and 209 $L$ turns, totaling 18k frames. Figure~\ref{fig:training} visualizes representative $U$ and $L$ demonstrations. All methods use this action-training set. Future prediction uses 25k frames with valid future targets from the same demonstration set. Appendix~\ref{app:data} specifies the sample selection criteria.

\paragraph{Baselines.}
We compare ACT~\citep{act}, Diffusion Policy~\citep{diffusion}, and a Base Flow Policy. ACT uses a conditional variational action decoder, and Diffusion Policy uses a transformer noise predictor with ten inference steps. Both use the same numeric, tactile, and global cube observations as the Base Flow Policy. The base flow model has the same action decoder as our method and a global point-set encoding. The ablation adds finger-relative geometry to this model, followed by the complete future-token prediction component. All models use a hidden width of 384, four observation-encoder layers, four action-decoder layers, a 20-step action horizon, and 1,000 training epochs. Detailed method-specific settings appear in Appendix~\ref{app:implementation}.

\paragraph{Layer-turn protocol.}
Each method is evaluated for three deployment rounds using one fixed checkpoint and deployment seeds 0, 1, and 2. Each round contains 100 attempts: 50 $U$ and 50 $L$, shuffled by the seed. A turn succeeds when the estimated target-angle error remains below $15^\circ$ for three observation frames. Each attempt has a ten-second timeout. A drop is detected when the cube moves more than 10 cm from its reference position in palm coordinates for three frames. After a timeout, the benchmark proceeds to the next scheduled attempt; dropped cubes are returned before continuation. We report the mean and sample standard deviation across the three rounds.

For $N$ attempts, mean time per attempt is
\begin{equation}
  T_{\mathrm{attempt}}=\frac{1}{N}\sum_{n=1}^{N}t_n,
  \label{eq:time}
\end{equation}
where $t_n$ is the elapsed time from policy execution start to success, timeout, or drop. Reset intervals between attempts are excluded. Appendix~\ref{app:benchmark} gives the results for each round.

\subsection{Comparison with Action-Learning Baselines}
\label{sec:baselines}
\begin{table}[t]
  \caption{\textbf{Real layer-turn comparison.} Success, timeout, and drop entries are percentages. Time is the mean duration per attempt in seconds. Values are mean $\pm$ sample standard deviation over three rounds of 100 attempts. Bold marks column-wise best values, including ties.}
  \label{tab:baseline}
  \centering
  \setlength{\tabcolsep}{3.2pt}
  \small
  \begin{tabular}{ccccccc}
    \toprule
    \thcell{Method} & \thcell{$U$ success $\uparrow$} & \thcell{$L$ success $\uparrow$} & \thcell{Timeout $\downarrow$} & \thcell{Drop $\downarrow$} & \thcell{All success $\uparrow$} & \thcell{Time $\downarrow$} \\
    \midrule
    \tableinput tables/baseline.tex
    \bottomrule
  \end{tabular}
\end{table}

Table~\ref{tab:baseline} shows that \ours{} achieves $\meanstd{99.0}{1.0}\%$ overall success. ACT, Diffusion Policy, and Base Flow achieve $77.7\%$, $78.7\%$, and $79.7\%$, respectively. The improvement over Base Flow is 19.3 percentage points. It appears in both directions: $U$ success rises from $82.7\%$ to $99.3\%$, and $L$ success rises from $76.7\%$ to $98.7\%$. The resulting policy supports consecutive turns of either layer under a common grasp.

The improvement comes mainly from fewer timeouts: three across 300 attempts, compared with 59 for Base Flow. Drops decrease from two to zero. Mean time per attempt is $\meanstd{5.25}{0.20}$ seconds, compared with $\meanstd{5.67}{0.14}$ for Base Flow, $\meanstd{6.06}{0.14}$ for Diffusion Policy, and $\meanstd{6.39}{0.62}$ for ACT.

\subsection{Representation Ablation}
\label{sec:ablation}
\begin{table}[t]
  \caption{\textbf{Representation ablation.} Local Geometry adds the finger-relative encoding to Base Flow. \ours{} additionally includes future tokens with the joint contact, progress, and motion prediction objective.}
  \label{tab:ablation}
  \centering
  \setlength{\tabcolsep}{3.2pt}
  \small
  \begin{tabular}{ccccccc}
    \toprule
    \thcell{Method} & \thcell{$U$ success $\uparrow$} & \thcell{$L$ success $\uparrow$} & \thcell{Timeout $\downarrow$} & \thcell{Drop $\downarrow$} & \thcell{All success $\uparrow$} & \thcell{Time $\downarrow$} \\
    \midrule
    \tableinput tables/ablation.tex
    \bottomrule
  \end{tabular}
\end{table}

Finger-relative geometry raises success to $\meanstd{92.7}{2.3}\%$, a gain of 13.0 percentage points over Base Flow (Table~\ref{tab:ablation}). Both directions reach $92.7\%$, while the timeout rate falls from $19.7\%$ to $7.0\%$. The geometry module uses the same reconstructed point set as the base policy and adds only its relation to each fingertip. These gains show that expressing cube geometry relative to each fingertip helps the policy complete turns more reliably.

Future-token prediction adds another 6.3 percentage points, increasing successful attempts from 278 to 297 out of 300. Timeouts fall from 21 to three, and drops from one to zero. Mean time per attempt remains close: 5.25 seconds for the full policy and 5.24 seconds for local geometry. Future prediction improves completion at a similar time cost by supplementing instantaneous geometry with supervision on layer progress, contact changes, and finger motion.

\subsection{Solving Real Scrambled Cubes}
\label{sec:solves}
\begin{figure}[!t]
  \centering
  \subfloat[$U$ turn\label{fig:real_U}]{\includegraphics[width=\linewidth]{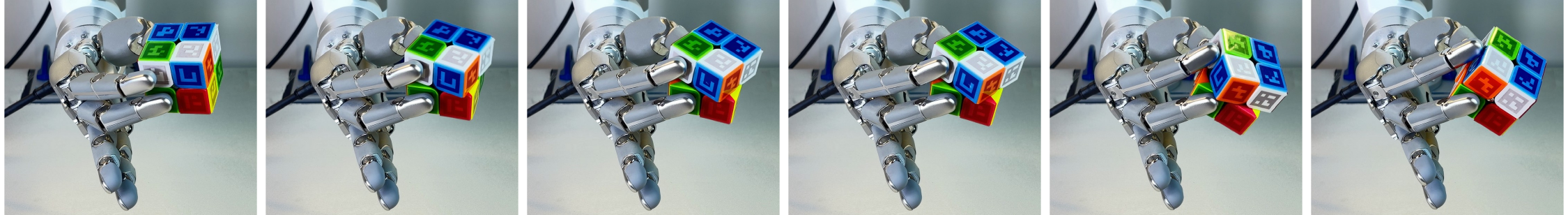}}
  \par\vspace{0.5em}
  \subfloat[$L$ turn\label{fig:real_L}]{\includegraphics[width=\linewidth]{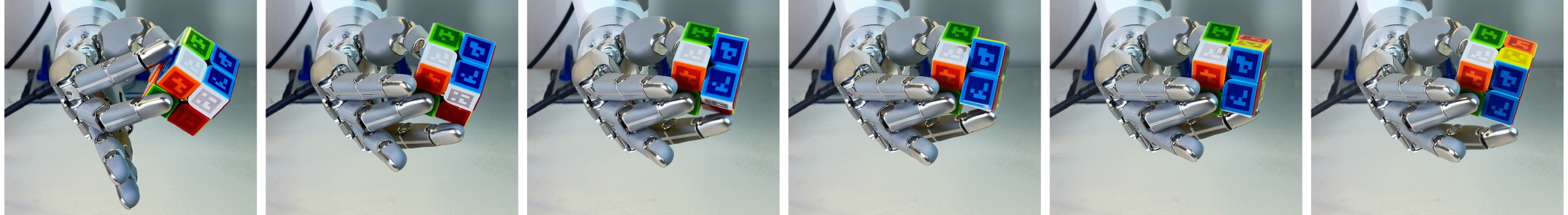}}
  \caption{\textbf{Real layer turns.} Six stages of each $90^\circ$ turn show layer advancement and finger reset while the supporting fingers retain the cube.}
  \label{fig:real_turns}
\end{figure}

\begin{figure}[!t]
  \centering
  \includegraphics[width=.9\linewidth]{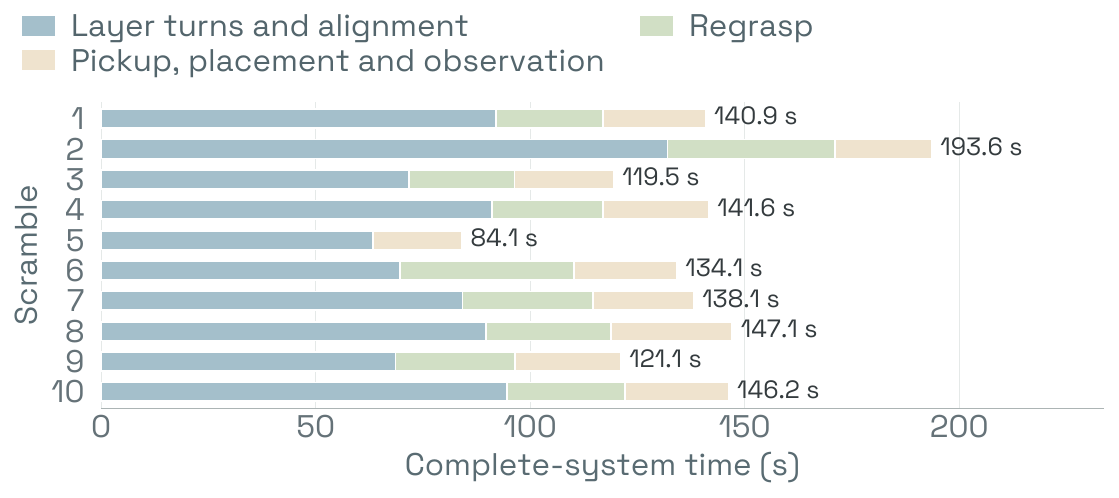}
  \caption{\textbf{Time spent in complete cube solving.} Layer turns dominate execution; nine solves include a regrasp. Bars cover the full system time, detailed in Table~\ref{tab:components}.}
  \label{fig:solve_timing}
\end{figure}

We evaluate ten random 11-move scrambles using $R$, $F$, and $U$ quarter, half, and inverse turns (Appendix~\ref{app:solves}). The solver uses the same layer-turn policy as the isolated benchmark. The complete solver uses an $18^\circ$ completion threshold for three observation frames and verifies the discrete state after each sequence.

\begin{table}[H]
  \caption{\textbf{Complete physical cube solves.} Counts and times cover planned $90^\circ$ attempts. Plan includes motion planning and checks; Total spans initial grasp planning through final placement and robot return. Times are seconds.}
  \label{tab:solves}
  \centering
  \small
  \setlength{\tabcolsep}{4.5pt}
  \begin{tabular}{ccccccccc}
    \toprule
    \thcell{Scramble} & \thcell{\#$U$} & \thcell{\#$L$} & \thcell{$U$ time $\downarrow$} & \thcell{$L$ time $\downarrow$} & \thcell{Time/$90^\circ$ $\downarrow$} & \thcell{Regrasps} & \thcell{Plan $\downarrow$} & \thcell{Total $\downarrow$} \\
    \midrule
    \tableinput tables/solves.tex
    \bottomrule
  \end{tabular}
\end{table}

\begin{figure}[H]
  \centering
  \includegraphics[width=.77\linewidth]{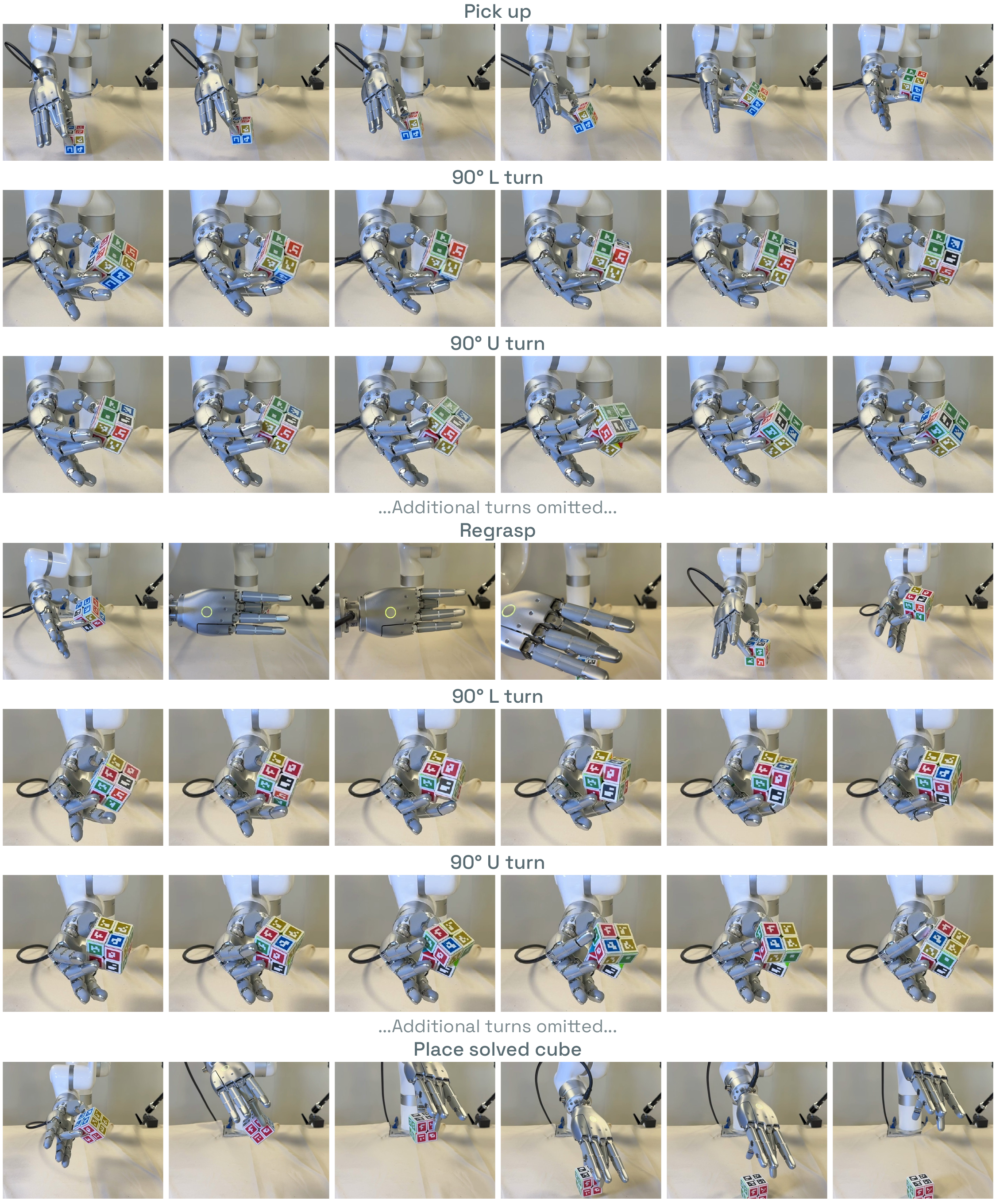}
  \caption{\textbf{From pickup to a solved cube.} Left to right, top to bottom. Each pair of turn rows shows the first two $90^\circ$ rotations after pickup or regrasp ($L$ then $U$); later turns are omitted.}
  \label{fig:real_solving}
\end{figure}

The system solves all ten cubes, achieving $10/10$ complete-solve success (Table~\ref{tab:solves}). Nine executions use one table-assisted regrasp; one completes under the initial grasp. Figure~\ref{fig:real_turns} shows physical $U$ and $L$ turns, and Figure~\ref{fig:real_solving} shows a complete execution. Complete-system time averages $\meanstd{136.63}{27.47}$ seconds, ranging from 84.08 to 193.59 seconds. Solutions require 12--26 planned quarter-turn attempts, with a per-solve average of $\meanstd{4.71}{1.32}$ seconds per quarter turn.

Across these solves, 181 of 185 planned turns succeed. Reobservation, alignment, and replanning recover from state mismatches to complete every solve. All four alignment corrections, ten initial pickups, nine regrasps, and ten final placements and returns succeed. State verification and recovery therefore allow the system to complete full solutions even when individual layer turns fail. Figure~\ref{fig:solve_timing} and Appendix~\ref{app:solves} break down its execution time.

\Needspace{4\baselineskip}
\section{Discussion and Conclusion}
\label{sec:discussion}
Continuous layer turns require a policy to coordinate each finger's spatial access to the target layer with changing contact and motion. Finger-relative geometry produces the largest improvement over the base policy, and future-token prediction further improves completion across both layers. Together, these components support repeated layer turns and integrate with pickup and table-assisted regrasping to solve scrambled cubes on real hardware.

The present evaluation uses one hand and a tagged $2\times2\times2$ cube, with marker-based reconstruction and planned pickup and placement. Extending the representation to unmarked articulated objects and other hands would test how broadly its spatial and temporal structure transfers. The ten complete solves demonstrate the integrated capability; a larger set of scrambles and starting grasps would characterize its operating range.

\clearpage
\subsection*{Acknowledgment}
This work was supported, in part, by NSF CAREER Award IIS-2240014 and NSF CCF-2112665 (TILOS).
\subsection*{AI use statement}
Generative AI assisted with literature retrieval, analysis code, figure preparation, and manuscript drafting and editing.
\subsection*{Reproducibility statement}
Appendix~\ref{app:implementation} specifies the observations, architecture, training objectives, and deployment settings. Appendix~\ref{app:benchmark} reports all evaluation rounds and timing definitions, and Appendix~\ref{app:solves} gives the complete scramble and solution sequences and system timing breakdown.
\bibliography{references}
\bibliographystyle{iclr2027_conference}
\clearpage
\appendix
\section{Implementation Details}
\label{app:implementation}
\subsection{Demonstrations and Training Data}
\label{app:data}
We use a motion-capture glove to teleoperate the index, ring, and little fingers. Geometric retargeting maps the operator's finger configurations to robot joint commands while the supporting fingers maintain their grasp. Hand state, tactile feedback, and cube geometry from multi-view reconstruction are synchronized at 10 Hz.

The demonstration set comprises 36 sequences, 576 turns, and 25,520 frames. Action training uses successful turns with at least three samples, valid cube poses and remaining angles within $10^\circ$ over the final three samples, and off-axis rotation within $10^\circ$ throughout. Each turn lasts at most eight seconds (80 samples). The resulting action dataset contains 18,113 frames with valid cube poses from 435 turns: 226 $U$ and 209 $L$ turns.

Future prediction uses all 36 demonstration sequences, including turns that do not meet the action-training criteria. Each current/future pair spans a single uninterrupted turn with valid cube poses throughout. Non-increasing timestamps and gaps greater than 0.15 seconds separate continuous segments. The prediction dataset contains 24,943 current frames with at least one valid future target, yielding 24,943, 22,635, and 19,750 pairs at offsets of one, five, and ten samples. We sample action and prediction batches separately.

\subsection{Observations and Encoders}
\label{app:encoder}
The five finger-state vectors use the order index, ring, little, thumb, and middle. Each has 24 entries: five joint-position slots, five velocity slots, five torque slots, six tactile force/torque channels, and three fingertip coordinates. Joint quantities are zero-padded for fingers with fewer than five joints. The index, ring, and little fingers contribute four, four, and five controlled joints, respectively. The remaining-angle input is measured in units of a $90^\circ$ turn.

The cube representation is constructed from eight tracked cubie centers. The reconstruction identifies the two four-cubie layers, then derives 24 face-center points and eight outer corners. Both these points and the fingertip positions are transformed into the palm frame. Numeric observations and global cube coordinates are standardized using training-set statistics and clipped to $[-5,5]$. Local geometry in Equation~\ref{eq:geometry} uses the unnormalized, unclipped palm-frame coordinates to preserve metric distances between the fingers and cube.

Each finger has a linear numeric projection to 384 dimensions. The global point network has dimensions $3\rightarrow384\rightarrow384$, with a GELU between its linear layers, followed by maximum pooling over the points. The local network has dimensions $3\rightarrow64\rightarrow64$, a GELU, mean pooling, and a $64\rightarrow384$ projection. The same local network is shared across fingers and points.

Tactile deformation images from the three controlled fingertips are resized to $96\times96$ and scaled to $[0,1]$. The shared convolutional encoder has channel widths 32, 64, 128, and 384, stride two at every layer, and kernels of size five in the first layer and three thereafter. A learned finger-specific scale and shift modulate every layer, followed by GELU activation. Global spatial averaging yields one token per tactile image. The observation transformer has four pre-normalized layers, eight attention heads, width 384, feed-forward width 1,536, zero dropout, and a final layer normalization. Its feed-forward sublayers use ReLU. The three learned future queries in the full model increase its memory from nine to 12 tokens. A shared $384\rightarrow384\rightarrow23$ MLP with GELU predicts the targets from each encoded future token.

\subsection{Action Normalization and Objective}
\label{app:objective}
Let $u_{t+k}\in\R^{13}$ denote a demonstrated commanded joint position and $q_t$ the measured configuration at the current observation. The action target is the offset $a_{t,k}=u_{t+k}-q_t$. For joint $j$, let $Q_{.01,j}$ and $Q_{.99,j}$ be the first and 99th percentiles of the training offsets. We use
\begin{equation}
    b_j=\frac{Q_{.01,j}+Q_{.99,j}}{2},\qquad
    s_j=\max\left(\frac{Q_{.99,j}-Q_{.01,j}}{2},0.01\right),\qquad
    A_{t,k,j}=\frac{a_{t,k,j}-b_j}{s_j},
    \label{eq:actionnorm}
\end{equation}
where the minimum scale is in radians. The action horizon is 20 samples. A binary mask $w_{t,k}$ excludes targets beyond the available turn segment. Writing $\bar s^2=\frac{1}{13}\sum_j s_j^2$, the optimized flow-policy loss is
\begin{equation}
    \mathcal L_{\mathrm{act}}=
    \frac{\sum_{t,k,j}w_{t,k}s_j^2(\widehat A_{t,k,j}-A_{t,k,j})^2}
    {13\bar s^2\sum_{t,k}w_{t,k}}.
    \label{eq:actloss}
\end{equation}
Thus the clean-action reconstruction error is measured in joint space and divided by the mean squared action scale. Noise levels are sampled from $\operatorname{Beta}(1.5,1)$. The velocity in Equation~\ref{eq:flow} is the parameterization used for integration; Equation~\ref{eq:actloss} is the training objective.

The four action-decoder blocks contain self-attention, cross-attention to the encoded observation memory, and a GELU feed-forward network. Each sublayer receives a noise-dependent shift and scale. A sinusoidal noise-level embedding uses logarithmically spaced periods from 0.004 to 4, followed by a two-layer SiLU MLP. Learned embeddings specify the 20 action positions. Four Euler updates with step size $-0.25$ evaluate the field at noise levels 1, 0.75, 0.5, and 0.25. The sampled normalized offsets are converted back with $b_j,s_j$ and added to the current measured joint configuration.

For future prediction, each target coordinate is divided by its empirical standard deviation for that horizon, with a minimum scale of $10^{-6}$. The standardization does not subtract a target mean. The contact target contains the three measured force components for each controlled finger; torque channels are observation inputs but are not part of this target. Equation~\ref{eq:predloss} averages only valid temporal pairs, with equal weight for the three target groups.

\subsection{Optimization and Baseline Settings}
\label{app:training}
We train each policy for 1,000 epochs (71,000 optimizer steps) with batches of 256 action examples. Training uses AdamW with $\beta=(0.9,0.95)$, weight decay $10^{-4}$, a 1,000-step learning-rate warmup followed by cosine decay, gradient-norm clipping at one, and bfloat16 mixed precision. Independent Gaussian noise with standard deviation 0.01 augments normalized numeric features and cube coordinates for action training. Exponential moving averages use decay 0.99 and provide the deployment parameters. All flow policies use learning rate $5\times10^{-4}$; ACT and Diffusion Policy use $10^{-4}$.

Our predictive branch draws an additional batch of 256 eligible current frames at each step while its weight is nonzero. The weight begins at 0.003 and decreases linearly to zero at step 35,500. Both action and prediction gradients update the same observation encoder. The action objective continues for the remaining training steps, retaining the learned future queries and the same action-conditioning memory.

ACT uses the shared observation interface with a 32-dimensional variational latent, a four-layer posterior transformer, an $L_1$ action reconstruction objective, and KL weight 10. The posterior reads the current joint configuration and the demonstrated chunk during training; inference fixes the latent to zero. Diffusion Policy uses 100 training diffusion steps with a cosine schedule, predicts noise with a four-layer transformer decoder, and uses ten deterministic denoising updates at deployment. Both methods have the same width, action horizon, and observation-encoder depth as Base Flow. Their input encoders use the global cube feature without the proposed local geometry or future-token prediction.

\subsection{Deployment and Discrete Planning}
Inference runs asynchronously from the 10 Hz command loop. An action sample is aligned to its observation timestamp, and execution selects the chunk entry corresponding to the current time. A newly inferred chunk can replace the active chunk after at least five executed samples; expired chunks can be replaced immediately. Commanded joint positions are clipped to robot joint limits. The benchmark uses the same execution rules for every method.

The discrete planner represents the orientations of the eight cubies and searches over positive $U/L$ quarter turns. Its objective first minimizes the number of quarter turns and then prefers lower repeated-layer cost among tied solutions. A compatible solved two-cubie block defines a holding grasp. If a complete solution is unavailable under the current grasp, the system creates such a block, places the cube, and picks it up in the new grasp before solving the remaining state. State observations between sequences determine whether alignment or another plan is needed.

\section{Complete Layer-Turn Results}
\label{app:benchmark}
For a round with $N=100$ attempts, the success, timeout, and drop percentages use all $N$ outcomes as the denominator. Direction-specific success uses the 50 attempts of that direction. For any per-round metric $x_r$, the reported summary is
\begin{equation}
    \bar x=\frac{1}{3}\sum_{r=1}^{3}x_r,\qquad
    s_x=\sqrt{\frac{1}{2}\sum_{r=1}^{3}(x_r-\bar x)^2}.
\end{equation}
The three rounds use deployment seeds 0, 1, and 2 with one checkpoint per method. Table~\ref{tab:rounds} gives the per-round results; Base Flow and \ours{} use the same evaluations in the baseline and ablation comparisons. For \ours{}, mean $U$ and $L$ times are $\meanstd{5.21}{0.32}$ and $\meanstd{5.29}{0.09}$ seconds, respectively, including failed attempts.

\begin{table}[t]
    \caption{\textbf{All layer-turn evaluation rounds.} Outcome entries are percentages. $U$, $L$, and All times average every attempt of the corresponding direction or the full round, including failures, in seconds. Each row contains 50 $U$ and 50 $L$ attempts.}
    \label{tab:rounds}
    \centering
    \footnotesize
    \setlength{\tabcolsep}{3pt}
    \begin{tabular}{cccccccccc}
        \toprule
        \thcell{Method} & \thcell{Seed} & \multicolumn{5}{c}{Outcomes (\%)} & \multicolumn{3}{c}{Mean time (s)} \\
        \cmidrule(lr){3-7}\cmidrule(lr){8-10}
         &  & \thcell{$U$ success $\uparrow$} & \thcell{$L$ success $\uparrow$} & \thcell{Timeout $\downarrow$} & \thcell{Drop $\downarrow$} & \thcell{All success $\uparrow$} & \thcell{$U$ $\downarrow$} & \thcell{$L$ $\downarrow$} & \thcell{All $\downarrow$} \\
        \midrule
        \tableinput tables/rounds.tex
        \bottomrule
    \end{tabular}
\end{table}

\section{Complete Cube-Solving Evaluation}
\label{app:solves}
\subsection{Scrambles, Solutions, and Corrections}
Table~\ref{tab:formulas} lists the ten scramble sequences and their corresponding solution plans. In a scramble, a prime denotes a negative quarter turn and a suffix 2 denotes a half turn in conventional cube notation. In a solution, $U$ and $L$ refer to the two controllable layers in the current robot grasp; suffixes 2 and 3 expand to two and three positive quarter turns. A slash marks the table-assisted regrasp between two solution sequences. The layer coordinate convention is therefore updated at that boundary. An arrow marks replanning under the same grasp.

\begin{table}[t]
    \caption{\textbf{Scramble and solution formulas.} Solution formulas use grasp-relative positive $U/L$ turns. The slash denotes a regrasp; an arrow denotes replanning under the same grasp. Small alignment corrections are timed separately.}
    \label{tab:formulas}
    \centering
    \small
    \setlength{\tabcolsep}{4pt}
    \begin{tabular}{ccc}
        \toprule
        \thcell{ID} & \thcell{Scramble} & \thcell{Planned solution sequences} \\
        \midrule
        \tableinput tables/formulas.tex
        \bottomrule
    \end{tabular}
\end{table}

The first solve requires a 1.80-second $L$ alignment after its first sequence. After the final planned $U$ turn in the third solve (6.30 seconds), a 0.70-second $U$ alignment completes the solution. The eighth and ninth solves use $L$ and $U$ alignment corrections of 2.30 and 3.20 seconds, respectively, before regrasping. These partial-angle corrections are reported separately from the planned $90^\circ$ turns. In solve 8, the post-regrasp plan stops after its twelfth attempt, a $U$ state mismatch; reobservation produces the plan \texttt{L3}. Its first $L$ attempt also produces a state mismatch, after which the next observation verifies a solved cube. Thus, solve 8 executes 20 planned quarter-turn attempts. Solve 9 similarly verifies completion after a state mismatch on its final planned $U$ turn.

\subsection{System Success and Timing}
\label{app:stages}
During complete cube solves, each turn continues until completion or a detected execution error. The ten-second per-attempt timeout applies to the isolated layer-turn benchmark.

Complete-system time spans initial grasp planning through final placement and robot return. For each solve, $U$ and $L$ times average all planned attempts of that direction. Time per $90^\circ$ is the total time of all such attempts divided by their count. We report the mean and sample standard deviation across ten solves. The full-solve success rate is $10/10$, and the planned-turn success rate is $181/185$. Across solves, mean $U$ and $L$ times are $\meanstd{5.10}{1.97}$ and $\meanstd{4.31}{0.86}$ seconds, respectively.

\begin{table}[t]
    \caption{\textbf{Time breakdown for each complete solve.} Seconds. Pickup includes initial planning, pregrasp, approach, closure, lift, and rotation into the manipulation pose. Regrasp includes intermediate placement and pickup in the new grasp. Observe includes the remaining state-observation and transition intervals.}
    \label{tab:components}
    \centering
    \small
    \setlength{\tabcolsep}{5pt}
    \begin{tabular}{cccccccc}
        \toprule
        \thcell{ID} & \thcell{Pickup $\downarrow$} & \thcell{Layer turns $\downarrow$} & \thcell{Alignment $\downarrow$} & \thcell{Regrasp $\downarrow$} & \thcell{Place/return $\downarrow$} & \thcell{Observe $\downarrow$} & \thcell{Total $\downarrow$} \\
        \midrule
        \tableinput tables/solve_components.tex
        \bottomrule
    \end{tabular}
\end{table}

Table~\ref{tab:stages} reports outcomes for each system stage. Motion planning and its associated checks (Table~\ref{tab:solves}) take $\meanstd{2.35}{1.04}$ seconds per solve and are included in the corresponding arm-motion stages.

\begin{table}[!htb]
    \caption{\textbf{System-stage outcomes and durations.} Success rates are percentages; time is mean $\pm$ sample standard deviation in seconds per operation. The first four stages form initial pickup.}
    \label{tab:stages}
    \centering
    \small
    \begin{tabular}{cccc}
        \toprule
        \thcell{Stage} & \thcell{Completed / attempted $\uparrow$} & \thcell{Success $\uparrow$} & \thcell{Duration $\downarrow$} \\
        \midrule
        \tableinput tables/stages.tex
        \bottomrule
    \end{tabular}
\end{table}

\end{document}